\documentclass[11pt]{article}
\usepackage[margin=1in]{geometry}
\usepackage[numbers,sort&compress]{natbib}
\usepackage[utf8]{inputenc}
\usepackage[T1]{fontenc}
\usepackage{amsmath,amssymb,booktabs,graphicx,microtype,url,xcolor}
\usepackage[hidelinks]{hyperref}
\usepackage{enumitem}
\usepackage{tikz}
\usetikzlibrary{arrows.meta,positioning,calc}
\graphicspath{{figures/}}
\newcommand{\clip}{\operatorname{clip}}
\newcommand{\ind}{\mathbf{1}}

\title{Cap-Aware Boosting for Green Day-Ahead\\ Electricity Price Forecasting}
\author{Shubham Pratap\\
AMPIN Energy Transition\\
\texttt{spratap@ampin.energy}}
\date{November 2026}

\begin{document}
\maketitle
\begin{abstract}
We forecast 15-minute Green Day Ahead Market (GDAM) prices with Prod v1: lagged GDAM/DAM/RTM prices, a LightGBM cap classifier, a log-price regressor, and nine quantile heads. On 257 delivery days (1 January--14 September 2026; 24{,}672 blocks) v1 has MAE 397.9 Rs/MWh and MAPE 9.57\%. That is 31.8\% lower MAE than a saved univariate Chronos-2 run (583.3) and 32.5\% lower than yesterday's profile (589.1). Prod v2, which adds auction fundamentals and recency weights, is close on aggregate MAE (395.5) with a 14-day block-bootstrap interval for v1 minus v2 of $[-7.8,12.6]$ Rs/MWh. From 25 October we compare the same question on a later, harder stretch. Exp5 adds morning real-time prices, yesterday's unmatched-sell shape, and a clear-sky solar curve on top of leak-safe Prod v2. Exp7 keeps those inputs, adds lagged wind and other renewable sell, and issues a block-dependent mix of quantiles instead of the cap blend. On 25--28 October, exp7 has the lowest MAE (789 Rs/MWh), ahead of exp5 (828), InputFix (1{,}019), and a saved slim Chronos-2 run (2{,}641). Over the 21 matched days through 16 November, exp5 is best (MAE 716) and Chronos-2 remains near 2{,}000.
\end{abstract}

\section{Introduction}
\label{sec:intro}
The Indian Energy Exchange Green Day Ahead Market (GDAM) enables renewable electricity transactions for next-day delivery through an auction process \citep{iexgdam}. Forecasting its market clearing price (MCP) requires resolving a full intraday price trajectory rather than a single daily average. A model can have a modest aggregate error while systematically missing morning changes, evening scarcity, or low-price daylight blocks. These differences matter when a price forecast is later used for scheduling or bidding, although forecast accuracy alone does not establish financial value.

The problem has two useful sources of structure. First, GDAM, the conventional day-ahead market (DAM), and the real-time market (RTM) provide related price histories. Their observations become available on different schedules and must be lagged accordingly. Second, a substantial fraction of prices in the evaluation period are close to the upper limit used by the deployed system. A model that predicts a smooth central price can underestimate these events. In the audited 2026 sample, 6,398 of 24,672 blocks have an actual price of at least 9,990 Rs/MWh. The remaining blocks span a wide range, with a minimum of 410.88 Rs/MWh.

Prod v1 blends a log-price LightGBM regressor with a cap-event classifier. Quantiles are fit separately and then shifted by a rolling residual correction. The system uses 49 price and calendar features, refits every 14 delivery days, and rebuilds each day's rows from history available before that day.

The 257-day comparison is against Prod v2, a saved Chronos-2 run, and two persistence baselines. V2 adds lagged auction fundamentals, short price trends, and exponential recency weights. Chronos-2 is a pretrained time-series model \citep{chronos2}; the archived run is univariate GDAM, no covariates, no fine-tune. That is the neural baseline in this paper.

The January--September comparison ends on 14 September. From 25 October onward we keep that question, but on a harder stretch of prices, and we add two later LightGBM variants. InputFix is leak-safe Prod v2. Exp5 adds morning real-time prices, yesterday's auction shape, and a clear-sky solar curve. Exp7 keeps those inputs, adds lagged wind and other renewable sell, and reads the forecast from a fixed mix of quantiles rather than from the cap blend. Section~\ref{sec:later} spells this out. The later Chronos-2 column is a saved slim checkpoint, not the univariate run in Table~\ref{tab:overall}.

\section{Related work}
\label{sec:related}
Electricity-price forecasting has a long history of statistical, machine-learning, and neural approaches. \citet{lago2021} emphasize robust baselines, sufficiently long evaluation periods, and statistically grounded comparisons. We follow those principles through a continuous 257-day evaluation, persistence baselines, and paired loss analysis, while recognizing that one market and an incomplete calendar year remain limited evidence.

Gradient-boosted trees fit this setting: mixed lag features, a hard cap, and a need for both a point forecast and quantiles. LightGBM supplies regression, classification, and quantile objectives in one library \citep{ke2017}. Tree ensembles remain competitive with neural models on many tabular tasks \citep{grinsztajn2022}. Chronos-2 is a pretrained time-series model that can take covariates \citep{chronos2}. The run we compare against does not: it is the released checkpoint, univariate GDAM, prediction length 96. That is a real production alternative, not a claim about every Chronos-2 configuration.

Quantile forecasting and prediction-interval calibration must also be distinguished. Conformalized quantile regression provides finite-sample guarantees under its stated assumptions \citep{romano2019}; adaptive conformal inference addresses changing distributions using a specific online update \citep{gibbs2021}. The residual-quantile correction in v1 implements neither procedure exactly. We therefore evaluate its empirical coverage and sharpness without borrowing guarantees from those methods.

\section{Forecasting task and information set}
\label{sec:task}
Let $y_{d,b}$ denote GDAM MCP for delivery date $d$ and block $b\in\{0,\ldots,95\}$. The forecast is $\hat y_{d,b}$ and signed deviation is $e_{d,b}=\hat y_{d,b}-y_{d,b}$. Prices and absolute errors are expressed in Rs/MWh. Dates and block times are interpreted in Indian Standard Time. The requested solar band is the fixed set $\mathcal S=\{24,\ldots,73\}$, corresponding to 06:00 inclusive to 18:30 exclusive. Its complement contains 46 blocks. These are reporting bands, not astronomical daylight estimates or labels for the generation technology of a transaction.

The evaluation grid is 1 January--14 September 2026: 257 consecutive dates and 24,672 unique date-block pairs per model. The three saved runs have no missing actuals or point forecasts on this grid, and their actual prices agree to numerical tolerance. We retain every block, including low prices and upper-price events. September contains only 14 days; none of the results is a full-year 2026 estimate. The exchange publishes GDAM market information at fifteen-minute resolution \citep{iexdata}.

\paragraph{Price features.}
For each market $m$, v1 constructs 13 predictors. GDAM and DAM use same-block lags of 1, 2, and 7 days. RTM uses lags of 2, 3, and 7 days, reflecting the unavailable portion of the immediately preceding RTM delivery day at a day-ahead decision point. Seven-day rolling means and standard deviations, a 28-day rolling median, and the difference between 7- and 28-day rolling means summarize local level and variability. Further features describe the most recent upper-price indicator, 7- and 28-day upper-price frequencies, and lagged daily mean, maximum, and upper-price count. All rolling operators act on already shifted matrices. Minimum valid counts are five for a 7-day window and twenty for a 28-day window.

There are 39 market features and 10 time/calendar features: block index, hour, intraday sine and cosine, day of week, weekend flag, month, annual sine and cosine, and Indian holiday flag. No target-day weather, auction fundamentals, or optional fresh-RTM features are enabled in the audited v1 run. Although the processed source contains many other columns, the v1 price loader reads only timestamp and MCP and reconstructs these 49 predictors.

\paragraph{Historical coverage.}
The processed price matrices used by v1 start on 1 May 2022; cleaned files used by v2 and Chronos-2 start on 1 April 2022. The 28-day-feature availability rule gives first usable training dates of 21 May and 21 April, respectively. At the first evaluation cutoff, v1 has 126,816 labeled training rows and v2 has 129,696. Chronos-2 receives 131,616 pre-2026 price observations before its internal context handling. Thus the evaluation targets are matched but the training histories differ. Calendar indices have no missing whole dates in the audited matrices, so positional daily shifts correspond to calendar-day lags. RTM retains missing cells that LightGBM can handle natively.

\section{The v1 forecasting system}
\label{sec:method}
\begin{figure}[ht]
\centering\includegraphics[width=\linewidth]{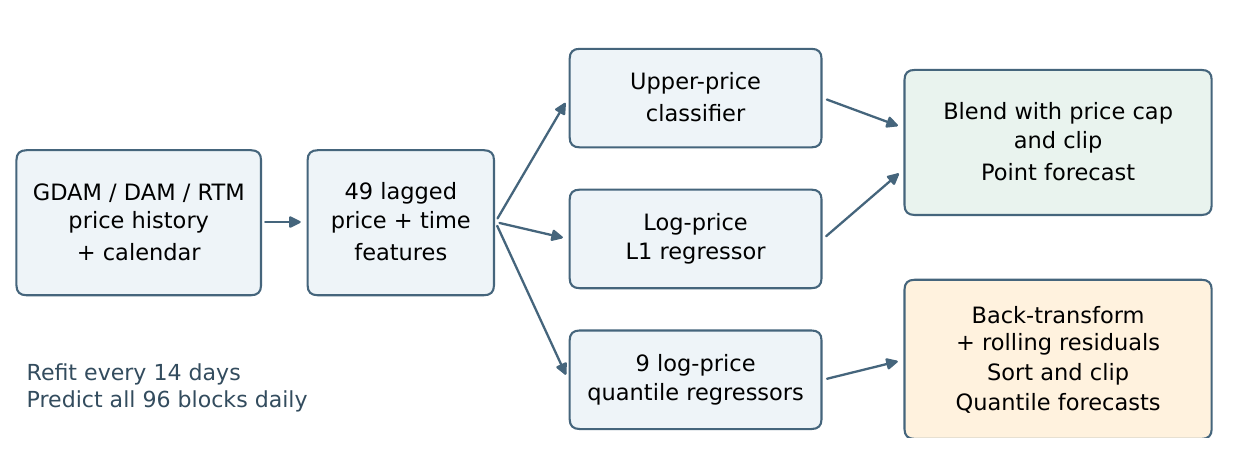}
\caption{V1's separate point and quantile pathways. The log-price point output is back-transformed before blending with the upper-price classifier. Rolling residual adjustments apply only to quantiles; all historical inputs precede the target date.}
\label{fig:system}
\end{figure}
\subsection{Point forecast and upper-price classifier}
For an input vector $x_i$, write $z_i=\log(1+\max(y_i,0))$. V1 fits an L1 regressor $f$ and a binary classifier $p$ using separate LightGBM ensembles:
\begin{align}
 f&\leftarrow\arg\min_f\sum_{i\in\mathcal T}|z_i-f(x_i)|,\label{eq:l1}\\
 c_i&=\ind\{y_i\geq 0.999C\},\qquad
 p\leftarrow\arg\min_p-\sum_{i\in\mathcal T}\big[c_i\log p(x_i)+(1-c_i)\log(1-p(x_i))\big],\label{eq:cap}\\
 \hat y(x)&=\clip_{[0,C]}\{p(x)C+[1-p(x)][\exp(f(x))-1]\},\qquad C=10{,}000.\label{eq:blend}
\end{align}
These optimization expressions describe the training losses, not a claim that finite boosting exactly attains their global minimizers. The L1 regressor is fitted on all retained rows, including upper-price observations. Consequently, Equation~\eqref{eq:blend} is a heuristic upward adjustment rather than a formally identified two-component mixture mean: its base predictor is not a conditional non-cap expectation. The point regressor and quantile regressors also do not share fitted trees.

The fixed threshold of 9,990 defines the model's upper-price regime. Historical training prices can exceed $C$; the processed pre-2026 maximum is 12,000 Rs/MWh and the cleaned history includes values up to 20,000. The classifier label therefore does not represent the contemporaneous regulatory cap throughout the historical training sample. We retain the original definition and treat time-varying caps as a needed extension. In the 2026 evaluation, actual prices lie in $[410.88,10{,}000]$, so the output clipping bounds contain all evaluated targets.

\subsection{Quantile forecasts and rolling residual adjustment}
For $q\in\mathcal Q=\{0.1,0.2,\ldots,0.9\}$, a separate regressor minimizes the pinball loss on log prices:
\begin{equation}
 g_q\leftarrow\arg\min_g\sum_{i\in\mathcal T}\rho_q(z_i-g(x_i)),\quad
 \rho_q(u)=\max(qu,(q-1)u).\label{eq:quantile}
\end{equation}
Back-transformed predictions are sorted within each block and clipped to $[0,C]$, producing $\tilde Q_{d,b}(q)$. Sorting enforces a nondecreasing quantile sequence; the upper-price point blend is not applied to these quantiles.

The first five evaluation days use these unadjusted quantiles. Thereafter the system takes the most recent 30 realized forecast days, or all available days if fewer, and computes
\begin{equation}
 \delta_{d,q}=\operatorname{Quantile}_{q}
 \left(\{y_{s,b}-Q_{s,b}(q):(s,b)\in\mathcal H_d\}\right),\qquad s<d.\label{eq:offset}
\end{equation}
It adds these offsets to $\tilde Q_{d,b}(q)$, sorts again, and clips. The stored $Q_{s,b}$ values include previous adjustments; the calibration history is therefore recursive. NumPy's default empirical-quantile interpolation is used. The point forecast remains unchanged apart from clipping. No assertion about improved point MAE can be attributed to this quantile-only correction.

This procedure is a rolling residual correction. Serial dependence, reuse of adjusted historical forecasts, and the lack of a finite-sample conformal order-statistic correction preclude an automatic distribution-free interpretation. A nominal 80\% interval is evaluated as $[Q(0.1),Q(0.9)]$; its observed coverage is an empirical diagnostic, not a coverage guarantee for future days.

\subsection{Training and daily execution}
The training set expands and contains only dates strictly before the current delivery date. The model bundle is refitted on the first day and every 14 days thereafter, yielding 19 bundles over the 257-day period. Between refits the feature rows incorporate new historical observations while fitted trees remain fixed. Every bundle comprises one point regressor, one classifier, and nine quantile regressors.

Each ensemble uses 350 estimators, learning rate 0.05, 63 leaves, minimum child count 40, row subsampling 0.9 at frequency 1, and column subsampling 0.8. Training uses available CPU threads. The source has no explicit seed setting for LightGBM; no variability across independently refitted seeds is estimated here. Appendix~\ref{app:repro} gives the implementation and reproduction details.

\section{Experimental protocol}
\label{sec:protocol}
\paragraph{Saved comparators.}
V2 uses the same model family and point rule, 97 features, and sample weights $w_i=2^{-a_i/180}$, where $a_i$ is training-row age in days relative to the latest training date. It adds lagged bids, market-cleared and cleared volumes, derived supply ratios, short price trends, and a level anchor. We evaluate its stored \texttt{forecast} column, not \texttt{forecast\_bc}. The comparison changes history length, covariates, and weighting simultaneously.

The Chronos-2 run uses AutoGluon 1.5.0, preset \texttt{chronos2}, checkpoint \texttt{autogluon/chronos-2}, prediction length 96, and no covariates. The log records seed 123, \texttt{skip\_model\_selection=True}, a 900-second budget every 14 days, and no fine-tune flag. Daily inputs are GDAM prices before the target date. We score the saved point and quantile columns as written \citep{autogluon}.

\paragraph{Persistence baselines.}
We additionally calculate previous-day and previous-week forecasts, $\hat y^{(1)}_{d,b}=y_{d-1,b}$ and $\hat y^{(7)}_{d,b}=y_{d-7,b}$, from the cleaned actual-price history. The required pre-January observations are available, so both baselines cover all 257 evaluation days. No parameters are fitted and no evaluation dates are dropped.

\paragraph{Metrics and paired uncertainty.}
We report MAE, root mean squared error (RMSE), mean signed deviation, MAPE, and weighted absolute percentage error (WAPE). For any evaluated set $\mathcal I$ of $N$ blocks,
\begin{equation}
 \operatorname{MAE}=\frac{1}{N}\sum_{i\in\mathcal I}|e_i|,\quad
 \operatorname{MAPE}=\frac{100}{N}\sum_{i\in\mathcal I}\frac{|e_i|}{\max(y_i,10^{-6})},\quad
 \operatorname{WAPE}=100\frac{\sum_{i\in\mathcal I}|e_i|}{\sum_{i\in\mathcal I}|y_i|}.\label{eq:metrics}
\end{equation}
The denominator floor is inactive because all evaluated prices are positive. MAPE nevertheless assigns higher weight to errors at lower actual prices \citep{hyndman2006}. Quantile loss is the mean original-price pinball loss across all nine quantiles. We also report nominal-80\% empirical coverage and mean interval width.

Paired comparisons use one mean loss per delivery day, preserving the 96 blocks within each day. For comparator $m$, $\Delta_d=L_d^{v1}-L_d^m$; negative values favor v1. We draw 10,000 circular block-bootstrap replicates of this daily series, with replacement, using 14-day blocks, and report 2.5th and 97.5th percentiles. Block lengths 7 and 28 provide sensitivity checks. Dependence-aware resampling follows the block-bootstrap principle \citep{kuensch1989}. These are approximate uncertainty intervals conditional on the saved forecasts; they neither measure refitting variability nor resolve test-set selection effects. Seasonal change limits the stationary-series approximation.

\section{What exp5 and exp7 add}
\label{sec:later}
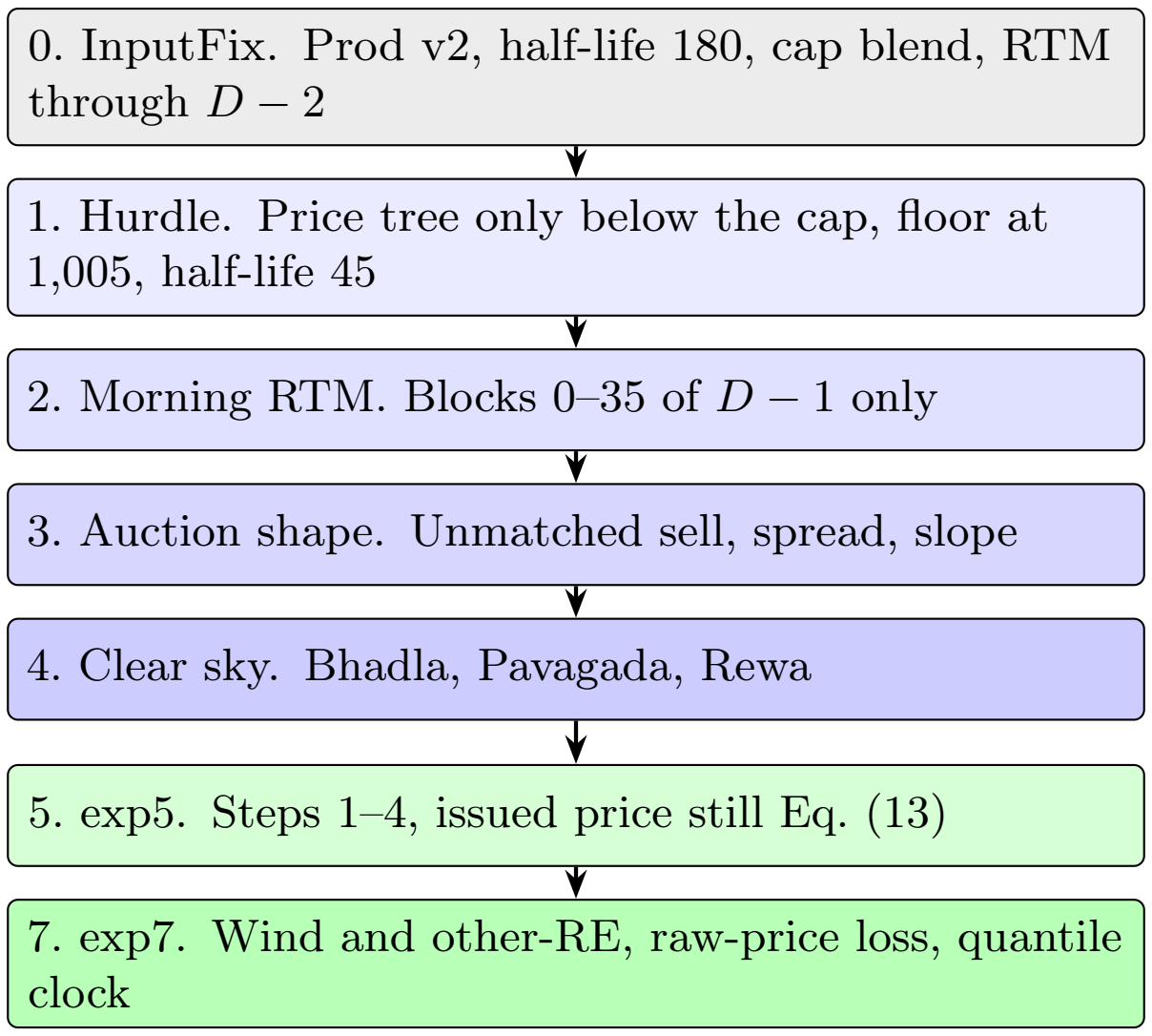
\begin{figure}[t]
\centering
\begin{tikzpicture}[
  font=\footnotesize,
  chg/.style={draw, rounded corners=2pt, align=left, inner sep=4pt, text width=7.6cm, minimum height=7mm},
  arr/.style={-{Stealth[length=2mm]}, thick}
]
\node[chg, fill=gray!15] (s0) {0.\ InputFix. Prod v2, half-life 180, cap blend, RTM through $D-2$};
\node[chg, fill=blue!8, below=2.2mm of s0] (s1) {1.\ Hurdle. Price tree only below the cap, floor at 1{,}005, half-life 45};
\node[chg, fill=blue!12, below=2.2mm of s1] (s2) {2.\ Morning RTM. Blocks 0--35 of $D-1$ only};
\node[chg, fill=blue!16, below=2.2mm of s2] (s3) {3.\ Auction shape. Unmatched sell, spread, slope};
\node[chg, fill=blue!20, below=2.2mm of s3] (s4) {4.\ Clear sky. Bhadla, Pavagada, Rewa};
\node[chg, fill=green!16, below=3mm of s4] (s5) {5.\ exp5. Steps 1--4, issued price still Eq.~\eqref{eq:exp5blend}};
\node[chg, fill=green!28, below=2.2mm of s5] (s7) {7.\ exp7. Wind and other-RE, raw-price loss, quantile clock};
\draw[arr] (s0) -- (s1);
\draw[arr] (s1) -- (s2);
\draw[arr] (s2) -- (s3);
\draw[arr] (s3) -- (s4);
\draw[arr] (s4) -- (s5);
\draw[arr] (s5) -- (s7);
\end{tikzpicture}
\caption{Each box keeps the box above it. Exp5 is the first full system we score. Exp7 is that system with a new loss and a new issued price.}
\label{fig:ladder}
\end{figure}

By September we already had a working day-ahead price: Prod v1, and then Prod v2. The later experiments do not throw that away. They start from InputFix, which is Prod v2 with one extra rule: the feature row for delivery day $D$ may use only what a desk can know at the 10:00 IST gate. From there the work is a stack. Each experiment keeps the previous one and adds one change. Table~\ref{tab:changes} is the list. Figure~\ref{fig:ladder} draws the same list. Exp5 is steps 1 through 4 trained together, and it still publishes the cap blend. Exp7 adds wind and other-renewable sell, changes the loss, and publishes the clock mix instead. Table~\ref{tab:exphyper} is every LightGBM setting that is shared and every setting that moves.

\begin{table}[t]
\centering\small
\caption{What is added at each step. Later rows keep everything above them. Exp5 is the first row we score as a full system. Exp7 is the last.}
\label{tab:changes}
\begin{tabular}{llp{7.2cm}}
\toprule
Step & Name & Change from the row above \\
\midrule
0 & InputFix & Prod v2. Half-life 180 days. Log-price tree on every row. Cap blend. RTM only through $D-2$. \\
1 & Hurdle & Price tree trained only below the cap. Floor classifier at 1{,}005. Half-life 45 days. Weight $(1000/y)^{0.35}$. \\
2 & Morning RTM & Blocks 0--35 of $D-1$ (00:00--08:45). Blocks 36--95 of that morning stay missing. \\
3 & Auction shape & Lagged unmatched sell, GDAM--DAM spread, neighbour blocks, slope. \\
4 & Clear sky & Three-hub GHI, west minus south, 7-day and 28-day drift. \\
5 & exp5 & Steps 1--4 together. Issued price is still the cap blend, Eq.~\eqref{eq:exp5blend}. \\
7 & exp7 & Wind and other-RE lags, thin-shelf flag. Point loss on raw price with weight $1/y$. Issued price is the clock mix, Eq.~\eqref{eq:prepeak}--\eqref{eq:eve}. \\
\bottomrule
\end{tabular}
\end{table}

\begin{table}[t]
\centering\small
\caption{LightGBM settings. Tree shape is shared. The rows that change between InputFix, exp5, and exp7 are the half-life, the point target, and the sample weight.}
\label{tab:exphyper}
\begin{tabular}{lccc}
\toprule
Setting & InputFix & exp5 & exp7 \\
\midrule
Trees & 350 & 350 & 350 \\
Learning rate & 0.05 & 0.05 & 0.05 \\
Leaves & 63 & 63 & 63 \\
Min.\ child samples & 40 & 40 & 40 \\
Row / column subsample & 0.9 / 0.8 & 0.9 / 0.8 & 0.9 / 0.8 \\
Seed & 42 & 42 & 42 \\
Quantile levels & 0.1--0.9 & 0.1--0.9 & 0.1--0.9 \\
Half-life (days) & 180 & 45 & 90 and 35 \\
Point target & $\log(1+y)$, all rows & $\log(1+y)$, $y<0.999C$ & raw $y$ \\
Extra weight & none & $(1000/y)^{0.35}$ & $1/y$ \\
Cap classifier & yes & yes & yes \\
Floor classifier ($y\le 1005$) & no & yes & no \\
Issued price & cap blend & cap blend & quantile clock \\
\bottomrule
\end{tabular}
\end{table}

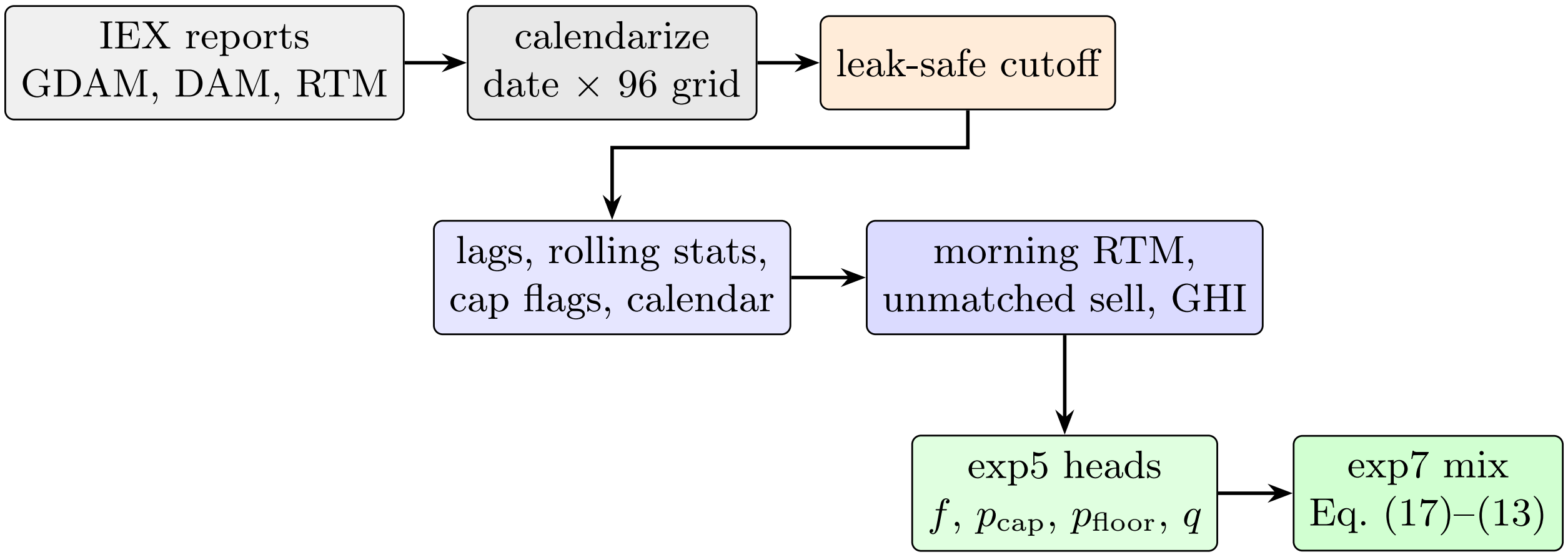
\begin{figure}[t]
\centering
\resizebox{\linewidth}{!}{%
\begin{tikzpicture}[
  font=\footnotesize,
  box/.style={draw, rounded corners=2pt, align=center, inner sep=3.5pt, minimum height=7.5mm},
  arr/.style={-{Stealth[length=2mm]}, thick}
]
\node[box, fill=gray!12] (raw) {IEX reports\\GDAM, DAM, RTM};
\node[box, fill=gray!18, right=5mm of raw] (cal) {calendarize\\date $\times$ 96 grid};
\node[box, fill=orange!15, right=5mm of cal] (cut) {leak-safe cutoff};
\node[box, fill=blue!10, below=8mm of cal] (lags) {lags, rolling stats,\\cap flags, calendar};
\node[box, fill=blue!14, right=6mm of lags] (extra) {morning RTM,\\unmatched sell, GHI};
\node[box, fill=green!12, below=8mm of extra] (e5) {exp5 heads\\$f$, $p_{\mathrm{cap}}$, $p_{\mathrm{floor}}$, $q$};
\node[box, fill=green!18, right=6mm of e5] (e7) {exp7 mix\\Eq.~\eqref{eq:prepeak}--\eqref{eq:exp5blend}};
\draw[arr] (raw) -- (cal);
\draw[arr] (cal) -- (cut);
\draw[arr] (cut.south) -- ++(0,-3mm) -| (lags.north);
\draw[arr] (lags) -- (extra);
\draw[arr] (extra) -- (e5);
\draw[arr] (e5) -- (e7);
\end{tikzpicture}}
\caption{How a delivery day is turned into a price. The top row is preprocessing. The bottom row is the model: exp5 fits the heads, exp7 writes the price from the quantile mix.}
\label{fig:exp_arch}
\end{figure}

The rest of this section walks the same stack in words, including the preprocessing cutoff and the formulas the two issued prices use. Figure~\ref{fig:exp_arch} is the data path those formulas sit on.

\subsection{First, make a day the desk could have seen}
Each IEX report is a list of time blocks. We turn GDAM, DAM, and RTM into the same grid: one row per calendar date, 96 columns for the fifteen-minute blocks, prices in Rs/MWh, plus purchase, sell, and cleared volume. If a date is missing we insert it. If a cell is missing we leave it missing. We do not copy yesterday into the hole, because that would look like a real price.

Delivery day $D$ is the day we are forecasting. Anything after the gate is deleted before a single lag is computed:
\begin{align}
\mathrm{GDAM},\ \mathrm{DAM}:&\quad \text{dates }\le D-1,\label{eq:cut_gdam}\\
\mathrm{RTM\ (InputFix)}:&\quad \text{dates }\le D-2,\label{eq:cut_rtm}\\
\mathrm{RTM\ (exp5 and exp7)}:&\quad \text{dates }\le D-1,\ \text{blocks }36\text{--}95\text{ of }D-1\text{ set to NaN}.\label{eq:cut_rtm_morn}
\end{align}
The third line is the only place exp5 is allowed to see ``today's morning.'' Blocks 0--35 of $D-1$ are 00:00--08:45 IST. By 10:00 they have already cleared. Blocks 36--95 of that same morning have not happened, so they stay blank. InputFix does not use even the early blocks; its RTM stops at $D-2$.

Lags, 7-day and 28-day means, and cap frequencies are shifts of this masked table, so they cannot see $D$. A cap flag is 1 when the price is at least $0.999C$ and stays missing when the price is missing. On Monday, yesterday is Sunday, so the workday lag uses the larger of Sunday and Saturday.

\subsection{Then add what the morning actually tells you}
With that table in hand we add three groups of columns. All of them are known before the gate.

The real-time market from the same morning. Night (blocks 0--19) contributes its mean and the change from the previous day. The ramp into the auction (blocks 24--35) contributes its mean, its maximum, the fraction of those blocks that sat on the cap, and how far that ramp sat from yesterday's DAM and GDAM on the same blocks.

Yesterday's auction, not just yesterday's price. Unmatched sell is sell bid minus cleared volume, lagged one day. When it is near zero the market usually cleared at 10{,}000. When a large volume is left unsold the price usually falls onto a buyer shelf. Alongside it we pass the GDAM--DAM spread, a flag for prices at or below 1{,}005 Rs/MWh, the neighbouring blocks of yesterday's GDAM curve, and the slope of that curve.

A solar clock with no weather feed. Clear-sky irradiance is computed at three hubs and mixed by installed-capacity weight:
\begin{equation}
\mathrm{GHI}_{\mathrm{India}}=0.55\,\mathrm{GHI}_{\mathrm{Bhadla}}+0.30\,\mathrm{GHI}_{\mathrm{Pavagada}}+0.15\,\mathrm{GHI}_{\mathrm{Rewa}}.\label{eq:ghi}
\end{equation}
Bhadla is $27.53^\circ$N, $71.91^\circ$E; Pavagada $14.10^\circ$N, $77.28^\circ$E; Rewa $24.48^\circ$N, $81.56^\circ$E. The model also sees west minus south, and how today's curve differs from 7 and 28 days ago. That difference is the drift of sunrise through the season.

\subsection{Exp5 writes the cap and the price as two jobs}
V1 asked one tree to learn the ordinary price and the 10{,}000 cap at the same time, then mixed them afterwards. Exp5 splits the job before training. Recent rows count more, and cheap blocks count a little more than a block already at the cap. With age $a_i$ in days and $C=10{,}000$,
\begin{align}
w_i&=2^{-a_i/45}\left(\frac{1000}{\max(y_i,100)}\right)^{0.35},\label{eq:exp5w}\\
c_i^{\mathrm{cap}}&=\ind\{y_i\ge 0.999C\},\qquad
c_i^{\mathrm{floor}}=\ind\{y_i\le 1005\}.\label{eq:exp5labels}
\end{align}
The log-price regressor $f$ is fit only where the price is below the cap. The cap classifier $p_{\mathrm{cap}}$ and a solar-floor classifier see every row. Nine quantile heads are still fit on log price, as in Equation~\eqref{eq:quantile}. When we publish a price we put the two jobs back together:
\begin{equation}
\hat y^{\mathrm{exp5}}
=\clip_{[0,C]}\Bigl\{p_{\mathrm{cap}}(x)\,C+\bigl(1-p_{\mathrm{cap}}(x)\bigr)\bigl[\exp(f(x))-1\bigr]\Bigr\}.\label{eq:exp5blend}
\end{equation}
If the classifier is sure of the cap, the price is 10{,}000. If it is sure there is no cap, the price is the uncapped tree. A shelf-snap flag is present in the code and is not used in the numbers below.

\subsection{Exp7 keeps those inputs and changes the issued price}
Exp5 still publishes one blend for every hour of the day. The errors were not like that. Midday wants the low quantiles. The hour before sunset wants a ramp from that low pair up toward the median and the upper quantile. Evening wants the cap probability back. Exp7 is that clock, written as formulas and then left fixed.

It trains on the same features, plus lagged wind sell and other-renewable sell (the level, the level against a 7-day mean, and for wind the 7-day mean over the 28-day mean) and a thin-shelf flag on unmatched sell. Two L1 models are fit on the raw price with weights $2^{-a_i/H}/y_i$, half-lives 90 and 35 days. The scored forecast is not those point models. It is a mix of the nine quantiles $q_1,\ldots,q_9$ (levels $0.1$ through $0.9$).

Away from the special hours the mix follows how high $q_7$ already sits:
\begin{align}
s&=\clip_{[0,1]}\frac{q_7-6000}{2800},\label{eq:capskew}\\
\hat y&=(1-s)\,(0.50 q_3+0.50 q_4)+s\,(0.65 q_1+0.35 q_2).\label{eq:defaultmix}
\end{align}
The clock then overwrites that value. From 07:15 to 08:45 the price is $0.50 q_2+0.50 q_3$. Through the solar day, 09:00 to 15:15, it sits on the floor,
\begin{equation}
\hat y_{\mathrm{solar}}=0.85 q_1+0.15 q_2.\label{eq:solar}
\end{equation}
The hour before the evening cap is the line we care about. On blocks 62--68 (15:30--17:00),
\begin{equation}
\begin{aligned}
\mathrm{prepeak}
&= (1-\mathrm{sunset\_pull})\,(0.50 q_2+0.50 q_3)\\
&\quad + \mathrm{sunset\_pull}\,(0.40 q_5+0.60 q_7),
\end{aligned}
\label{eq:prepeak}
\end{equation}
with $\mathrm{sunset\_pull}=0$ on blocks 62--65 and $1$ from block 66. Early in that window the forecast is still the low pair. At the end of the window it has moved to $0.40 q_5+0.60 q_7$. After 17:15 the cap classifier returns:
\begin{equation}
\hat y_{\mathrm{eve}}=
\begin{cases}
C & p_{\mathrm{cap}}\ge 0.60,\\
p_{\mathrm{cap}}\,C+(1-p_{\mathrm{cap}})(0.40 q_2+0.60 q_3) & \text{otherwise.}
\end{cases}
\label{eq:eve}
\end{equation}
Sunday evening and Monday use their own mixes. Those coefficients were chosen on earlier misses and were not refit on the October--November table.

The Chronos-2 column in the later tables is a separate saved run, the slim checkpoint, with history cut at the previous midnight. It is there as a comparison. It is not part of the exp5 or exp7 training.

\section{Results}
\label{sec:results}
\subsection{Aggregate forecasting performance}
\begin{table}[ht]
\centering\small
\caption{Point-forecast results on all 24,672 blocks, 1 January--14 September 2026. MAE, RMSE, and bias are in Rs/MWh; MAPE and WAPE are percentages. Bias is forecast minus actual.}
\label{tab:overall}
\begin{tabular}{lrrrrr}
\toprule
Model & MAE & RMSE & MAPE (\%) & WAPE (\%) & Bias \\
\midrule
Prod v1 & 397.9 & 849.1 & 9.573 & 7.677 & -6.1 \\
Prod v2 & 395.5 & 855.6 & 9.628 & 7.630 & 4.4 \\
Chronos-2 & 583.3 & 1,196.4 & 13.749 & 11.254 & 33.7 \\
Previous day & 589.1 & 1,281.6 & 14.037 & 11.367 & -13.4 \\
Previous week & 1,241.0 & 2,194.9 & 27.844 & 23.945 & -90.8 \\
\bottomrule
\end{tabular}
\end{table}

V1 MAE is 397.9 Rs/MWh against 583.3 for Chronos-2 and 589.1 for yesterday's profile (31.8\% and 32.5\% reductions). RMSE follows the same order: 849.1 versus 1,196.4 for Chronos-2. Chronos-2 here is the univariate pretrained run described above.

V2 has the smallest MAE, 395.464 Rs/MWh, a 0.61\% improvement over v1. The ranking reverses for MAPE (9.573\% for v1 versus 9.628\% for v2) and RMSE (849.1 versus 855.6). V1 has lower daily MAE than v2 on 136 of 257 days. Thus a single rounded aggregate score conceals differences in both daily frequency of wins and tail behavior.

\begin{table}[ht]
\centering\small
\caption{Paired daily MAE differences (v1 minus comparator), 95\% circular block-bootstrap intervals with 14-day blocks, and days with strictly lower v1 MAE. Differences and bounds are in Rs/MWh; negative differences favor v1.}
\label{tab:bootstrap}
\begin{tabular}{lrrr}
\toprule
Comparator & $\overline\Delta$ & 95\% interval & v1 wins / 257 \\
\midrule
Prod v2 & 2.4 & [-7.8, 12.6] & 136 \\
Chronos-2 & -185.4 & [-250.4, -127.7] & 202 \\
Previous day & -191.2 & [-240.1, -144.9] & 196 \\
Previous week & -843.1 & [-1,086.5, -620.5] & 236 \\
\bottomrule
\end{tabular}
\end{table}

The 14-day bootstrap interval for v1 minus Chronos-2 is $[-250.4,-127.7]$ Rs/MWh, and that against previous-day persistence is $[-240.1,-144.9]$. Both remain below zero with 7- and 28-day blocks. The interval against v2 is $[-7.8,12.6]$, which spans zero for every tested block length. This is insufficient evidence of a stable aggregate MAE distinction between v1 and v2 under this resampling analysis; it is not a formal equivalence result.

\subsection{Intraday and seasonal structure}
\begin{figure}[t]
\centering\includegraphics[width=\linewidth]{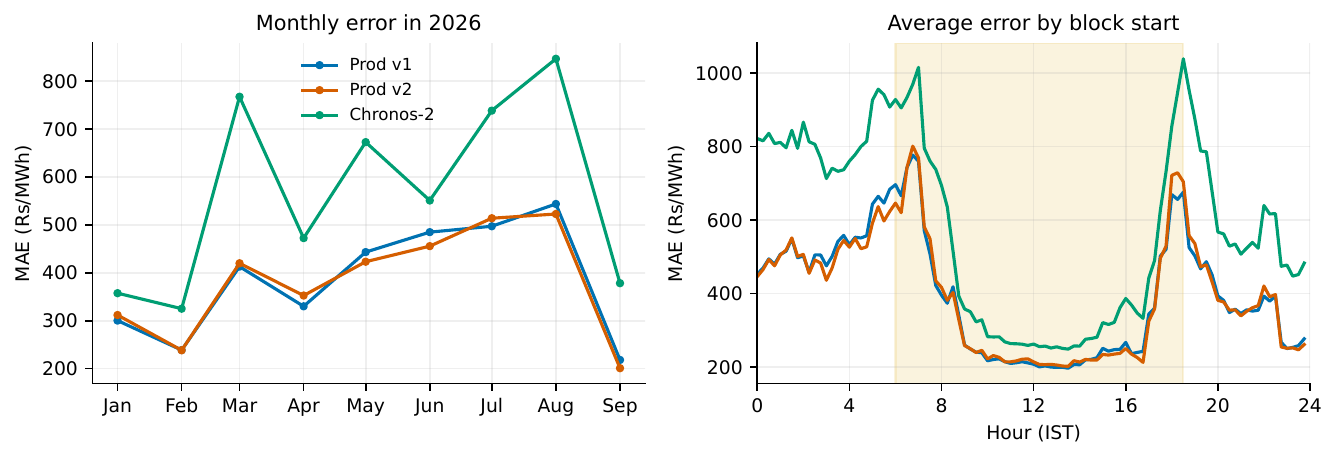}
\caption{Monthly MAE and average MAE by fifteen-minute block in the matched 2026 evaluation. September covers days 1--14 only. Shading denotes the fixed 06:00--18:30 solar reporting band.}
\label{fig:profiles}
\end{figure}
\begin{table}[ht]
\centering\small
\caption{Solar (06:00--18:30, 12,850 blocks) and non-solar (11,822 blocks) errors. Absolute errors are in Rs/MWh. These are fixed reporting bands.}
\label{tab:solar}
\begin{tabular}{lrrrr}
\toprule
Model & Solar MAE & Non-solar MAE & Solar MAPE (\%) & Non-solar MAPE (\%) \\
\midrule
Prod v1 & 337.7 & 463.3 & 11.474 & 7.507 \\
Prod v2 & 339.7 & 456.1 & 11.667 & 7.412 \\
Chronos-2 & 455.6 & 722.0 & 15.125 & 12.254 \\
\bottomrule
\end{tabular}
\end{table}

V1's solar MAE is 337.7 Rs/MWh and its non-solar MAE is 463.3. Relative to v2, it gains 2.0 Rs/MWh in the solar band but loses 7.2 in the non-solar band. Its improvements over Chronos-2 are 117.9 and 258.7 Rs/MWh, respectively. The larger absolute non-solar improvement suggests that the aggregate comparison is not driven only by low-price daylight periods. The equal-time weighting is explicit: 50 solar and 46 non-solar blocks contribute to each day's overall score.

Solar MAPE is higher than non-solar MAPE for v1 despite lower solar MAE. This reflects the smaller denominators associated with lower solar-hour prices; it does not contradict the MAE result. The original-price and percentage errors answer different questions. Figure~\ref{fig:profiles} also shows temporal heterogeneity. V1's complete-month MAE ranges from 238.7 in February to 543.4 in August. The 218.1 figure for the incomplete September period should not be interpreted as a comparable full-month result.

\subsection{Uncertainty quality and difficult regimes}
\begin{table}[ht]
\centering\small
\caption{Postprocessed quantile forecasts: mean pinball loss over nine quantiles, empirical coverage and mean width of the nominal-80\% interval, and fraction of point forecasts outside that interval. Loss and width are in Rs/MWh.}
\label{tab:uncertainty}
\begin{tabular}{lrrrr}
\toprule
Model & Pinball loss & Coverage (\%) & Width & Point outside (\%) \\
\midrule
Prod v1 & 171.6 & 81.21 & 1,394.6 & 0.31 \\
Prod v2 & 192.0 & 83.22 & 3,565.6 & 0.48 \\
Chronos-2 & 234.2 & 78.30 & 1,595.9 & 0.07 \\
\bottomrule
\end{tabular}
\end{table}

V1 has the lowest mean pinball loss among the three saved probabilistic forecasts. Its nominal-80\% interval covers 81.21\% of observations with a mean width of 1,394.6 Rs/MWh. V2 attains 83.22\% coverage but with mean width 3,565.6, about 2.56 times v1's. Chronos-2 achieves 78.30\% with width 1,595.9. Coverage alone therefore gives an incomplete assessment: wider intervals can attain higher coverage while being less informative. All saved quantile sequences are monotone after postprocessing.

\begin{figure}[t]
\centering\includegraphics[width=\linewidth]{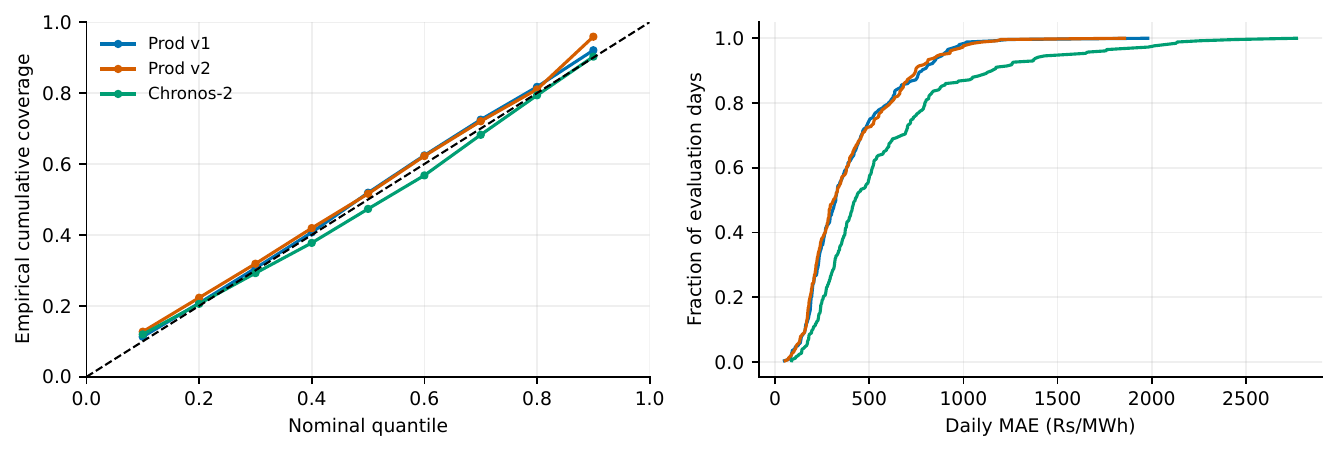}
\caption{Left: empirical cumulative quantile coverage; the diagonal is the continuous-outcome calibration reference, with deviations also affected by ties at the clipping boundary. Right: empirical distribution of daily MAE. Each model contributes the same 257 days.}
\label{fig:uncertainty}
\end{figure}
Aggregate coverage masks failure modes. On the 464 blocks below 1,000 Rs/MWh, v1's nominal-80\% coverage is only 64.01\%, despite its 81.21\% overall value. V1's absolute bias is also price dependent: mean deviation is positive in this low-price subset and negative on upper-price events. On blocks at or above 9,990, v1 MAE is 374.4, v2 MAE is 344.9, and Chronos-2 MAE is 562.9 Rs/MWh. These subsets are defined using realized prices and are diagnostics, not information available when issuing a forecast.

The upper tail remains material: v1's 99th-percentile absolute block error is 4,135.1 Rs/MWh, despite a median absolute error of 153.5. Near-zero mean bias does not mean cheap tails. The point forecast also sits outside the central 80\% band on 0.31\% of v1 blocks.

\subsection{October and November}
\begin{table}[ht]
\centering\small
\caption{Later holdout, 25--28 October 2026 (4 days, 384 blocks). Chronos-2 here is the saved slim AutoGluon checkpoint \texttt{models\_gdam\_chronos2} (leak-safe history before each day), not the univariate 257-day run in Table~\ref{tab:overall}. MAE is Rs/MWh.}
\label{tab:later}
\begin{tabular}{lrrr}
\toprule
Model & MAE & MAPE (\%) & Solar MAPE (\%) \\
\midrule
Chronos-2 (slim) & 2641.1 & 109.95 & 171.77 \\
InputFix Prod v2 & 1018.5 & 23.48 & 20.68 \\
exp5 (\texttt{v3\_full}) & 827.7 & 19.20 & 15.86 \\
exp7 (\texttt{seasonal\_master}) & 789.2 & 16.69 & 15.05 \\
\bottomrule
\end{tabular}
\end{table}

\begin{table}[ht]
\centering\small
\caption{Extended later window, 25 October--16 November 2026. Matched 21 delivery days (2{,}016 blocks) with Chronos-2, exp5, and exp7. 17 November is forecast-only (no actual MCP yet) and is omitted.}
\label{tab:later_long}
\begin{tabular}{lrrr}
\toprule
Model & MAE & MAPE (\%) & Solar MAPE (\%) \\
\midrule
Chronos-2 (slim) & 2000.4 & 108.67 & 182.19 \\
exp7 (\texttt{seasonal\_master}) & 823.8 & 20.80 & 25.27 \\
exp5 (\texttt{v3\_full}) & 716.0 & 19.07 & 21.90 \\
\bottomrule
\end{tabular}
\end{table}

\begin{table}[ht]
\centering\small
\caption{November overlap with InputFix Prod v2 (10 days with block-level InputFix files: 6--12 and 14--16 Nov). Same Chronos-2 and LightGBM stacks as above.}
\label{tab:later_nov}
\begin{tabular}{lrrr}
\toprule
Model & MAE & MAPE (\%) & Solar MAPE (\%) \\
\midrule
Chronos-2 (slim) & 1417.7 & 97.89 & 168.23 \\
InputFix Prod v2 & 708.1 & 26.12 & 35.68 \\
exp7 (\texttt{seasonal\_master}) & 696.7 & 21.69 & 30.46 \\
exp5 (\texttt{v3\_full}) & 606.7 & 20.27 & 27.51 \\
\bottomrule
\end{tabular}
\end{table}

Table~\ref{tab:later} is the first four days, 25--28 October. Chronos-2 is not close: MAE 2{,}641 Rs/MWh and solar MAPE 172\%, against 1{,}019 and 21\% for InputFix. Exp5 brings MAE down to 828 and exp7 to 789. On this short cut the Monday-aware quantile mix in exp7 is the better of the two LightGBM variants.

The picture changes once the window is long enough to include early November. Table~\ref{tab:later_long} matches Chronos-2, exp5, and exp7 on 21 delivery days through 16 November (2{,}016 blocks). Exp5 is then the best of the three, MAE 716 and MAPE 19.1\%. Exp7 is 824 / 20.8\%. Chronos-2 remains near 2{,}000 / 109\%. The extra quantile rules in exp7 helped on the late-October days and did not help across the full three weeks.

InputFix block files are not stored for every one of those 21 days. On the 10 November days where they are (6--12 and 14--16 November; Table~\ref{tab:later_nov}), the order is the same: exp5 607, exp7 697, InputFix 708, Chronos-2 1{,}418. Solar-hour MAPE moves with MAE. None of these October--November numbers are mixed into the 257-day table.

\section{Limitations}
\label{sec:limits}
The 257-day table is one market and one saved run of each system. V2's features were chosen with July 2026 diagnostics in hand, so January--September is a retrospective check, not an untouched holdout. Exp5 and exp7, and the slim Chronos-2 checkpoint they are compared with, cover 25 October--16 November only. The quantile mix inside exp7 was fixed from earlier misses; it was not re-tuned on the 21-day table, but it is also not a pre-registered rule.

V2's extra history, recency weights, and auction features move together. Solar/Non-Solar bid fields in the local files go missing on 8 July 2026 when the exchange renamed them to Wind/Other-RE/DRE, so v2 does not have a complete renewable mix through the main window. The quantile residual shift has no coverage theorem; coverage on prices below 1{,}000 Rs/MWh is 64\%. Logged wall-clock times (v1 17.2 min, v2 38.0, Chronos-2 11.2) include I/O and are not a speed benchmark. This paper does not evaluate bids or trading P\&L.

\section{Conclusion}
\label{sec:conclusion}
On the 257 delivery days through 14 September, Prod v1 cuts MAE by about a third relative to the saved univariate Chronos-2 run and to yesterday's price (397.9 against 583.3 and 589.1 Rs/MWh). V2 is tied with v1 on that aggregate. On the later October--November days the ranking among the boosting models is tighter: exp7 is best on the first four days, and exp5 is best once the window runs through 16 November (MAE 716 against 824 for exp7, about 700 for InputFix where we have it, and about 2{,}000 for the slim Chronos-2 checkpoint). The gain is in the inputs---morning RTM, yesterday's unmatched sell, and the clear-sky curve---not in replacing the tree model with Chronos-2.

\clearpage
\bibliographystyle{plainnat}
\bibliography{references}

\clearpage
\appendix
\section{Implementation and reproduction}
\label{app:repro}
\subsection{Pinned evaluation artifacts}
The empirical analysis uses the following combined prediction files under \texttt{pipeline\_state\_retro}. Their SHA-256 digests, sizes, exact paths, and the source snapshots are recorded in the accompanying analysis manifest.
\begin{itemize}[leftmargin=*,itemsep=2pt]
\item V1: \path{gdam_wf_2026-01-01_2026-09-14_20260914_135652/combined_predictions.csv}.
\item V2: \path{gdam_v2_hl180_20260914_141400/combined_predictions.csv}.
\item Chronos-2: \path{gdam_chronos2_15min_20260914_145209/combined_predictions.csv}.
\end{itemize}
The two persistence baselines use \path{iex_data_processed/markets/gdam_cleaned_15min.csv}. Their test-period actuals agree with all three saved prediction files. The analysis script asserts an exact 257-by-96 grid, unique keys, integer block indices in $[0,95]$, finite actuals and point forecasts, and actual-price equality with absolute tolerance $10^{-8}$. No missing model forecasts are filled, interpolated, or substituted.

\subsection{Feature availability and initial sample sizes}
\begin{table}[ht]
\centering\small
\caption{History available at the first evaluation cutoff, 1 January 2026. Tree-model rows follow feature warm-up. Chronos-2 rows denote the input series before the model truncates context.}
\label{tab:history}
\begin{tabular}{lllrl}
\toprule
Model & History begins & Usable rows begin & Rows & Inputs \\
\midrule
Prod v1 & 1 May 2022 & 21 May 2022 & 126,816 & 49 features \\
Prod v2 & 1 April 2022 & 21 April 2022 & 129,696 & 97 features \\
Chronos-2 & 1 April 2022 & 1 April 2022 & 131,616 & Univariate \\
\bottomrule
\end{tabular}
\end{table}

Historical matrices are indexed by delivery day, with columns representing the 96 blocks. The audit finds no missing whole days in either the processed or cleaned market indices. Their overlapping prices are equal. The RTM matrices have 171 missing cells over their full available ranges; the implementation uses window minimum counts and native missing-value handling rather than imputing a price from the future. V1's first usable row follows twenty available lagged daily observations, which is why the 28-day summary does not impose a full 28-day warm-up.

The complete 49- and 97-column feature schemas are saved in \texttt{analysis/feature\_audit.json}. Features exclude \texttt{date} and target \texttt{MCP}. Holiday encoding is active when the holidays package is available and otherwise falls back to zero; the inspected environment contains that package. The audited v1 run enables neither weather nor the optional fundamentals and fresh-RTM paths. Thus the presence of those functions in the repository is not evidence that they influenced these results.

\subsection{Execution details}
\begin{table}[ht]
\centering\small
\caption{Tree-model settings read from the audited implementation.}
\label{tab:hyperparameters}
\begin{tabular}{ll}
\toprule
Setting & Value \\
\midrule
Estimators per ensemble & 350 \\
Learning rate / leaves & 0.05 / 63 \\
Minimum samples per leaf & 40 \\
Row subsampling / frequency & 0.9 / 1 \\
Column subsampling & 0.8 \\
Objectives & L1 point; binary event; quantile \\
Quantile levels & 0.1 through 0.9 in steps of 0.1 \\
Price bound / event threshold & 10,000 / 9,990 Rs/MWh \\
Refit spacing / bundles & 14 days / 19 \\
Residual-adjustment history & Up to 30 days; starts after 5 days \\
v1 weights / v2 half-life & Uniform / 180 days \\
LightGBM seed in runner & Not explicitly set \\
\bottomrule
\end{tabular}
\end{table}

V1 and v2 rebuild their feature rows daily through the same fixed lag definitions, and fit on all retained dates before the refit origin. The refit calendar is 1 January and every fourteen days through 10 September, for nineteen refits. A bundle contains eleven separately fitted LightGBM ensembles. No early-stopping callback or holdout-based hyperparameter search is passed by these runners. V2's optional bias-corrected point column is saved alongside the principal forecast, but is excluded from this comparison.

The archived Chronos log records AutoGluon 1.5.0, Python 3.10.8, PyTorch 2.7.1+cu118, CUDA 11.8, 22 reported CPUs, one 8-GB GPU, and approximately 63.4 GB system RAM. It names the single \texttt{Chronos2} model and the checkpoint path \texttt{autogluon/chronos-2}. The configuration uses no exogenous covariates and no fine-tuning flag. The model implementation defaults to a finite context and uses pretrained inference; full earlier history passed to the wrapper does not mean that every historical point enters each attention computation. The exact immutable checkpoint revision is not present in the run log.

The inspected Python 3.12 analysis environment contains LightGBM 4.7.0, pandas 3.0.3, NumPy 2.2.6, and holidays 0.101. These versions document the audit environment; without an original lockfile they are not asserted to be the exact historical training environment. The accompanying analysis requires only pandas, NumPy, and Matplotlib for recomputing metrics and figures; it does not refit any model. Its random generator is NumPy \texttt{default\_rng} with seed 20260915.

\subsection{Reproduction commands and release status}
The recorded training commands are shown for reproducibility; rerunning them is a new experiment and can write new state or replace shared intermediate outputs.
\begin{verbatim}
python retro_daily_forecast_pipeline.py --market gdam \
  --start 2026-01-01 --end 2026-09-14 --retrain-every 14
python retro_daily_forecast_pipeline_v2.py --market gdam \
  --start 2026-01-01 --end 2026-09-14 --features v2 \
  --skip-verify --retrain-every 14
python retro_chronos2_gdam.py --start 2026-01-01 \
  --end 2026-09-14 --retrain-every 14 --time-limit 900
\end{verbatim}
The stored v2 run bypassed its built-in actual-price check; the independent comparison in this paper reinstates a stricter check for every matched block. To regenerate this paper's tables from existing forecasts, run \texttt{analyze\_results.py}; \texttt{audit\_features.py} reconstructs the input schemas without fitting. The project-root path is explicit in both scripts and should be adjusted for a different local installation.

This package documents the analysis and source files for author review. It does not establish public availability or permission to redistribute the underlying IEX data or employer-owned code. Source ownership, dataset terms, and checkpoint licensing need to be confirmed before an external release. Code hashes identify what was inspected, but they cannot establish that a later source snapshot is byte-identical to the code executed on 14 September.

\section{Additional diagnostics}
\label{app:diagnostics}
\subsection{Monthly and daily results}
\begin{table}[ht]
\centering\small
\caption{Monthly results in 2026. The first three numeric columns are MAE (Rs/MWh); the last three are MAPE (\%). September is partial.}
\label{tab:monthly}
\begin{tabular}{lrrrrrr}
\toprule
Month & v1 MAE & v2 MAE & C-2 MAE & v1 MAPE & v2 MAPE & C-2 MAPE \\
\midrule
Jan & 300.2 & 311.8 & 357.3 & 7.08 & 7.28 & 7.96 \\
Feb & 238.7 & 238.3 & 325.0 & 5.85 & 5.72 & 7.27 \\
Mar & 412.7 & 419.8 & 767.1 & 8.43 & 8.68 & 14.98 \\
Apr & 329.9 & 352.5 & 472.2 & 8.53 & 9.06 & 12.09 \\
May & 443.1 & 423.0 & 672.4 & 10.71 & 10.36 & 17.16 \\
Jun & 484.7 & 455.4 & 550.7 & 11.05 & 11.75 & 12.93 \\
Jul & 496.9 & 513.7 & 738.1 & 13.01 & 13.34 & 18.55 \\
Aug & 543.4 & 522.5 & 846.2 & 13.74 & 12.95 & 21.27 \\
Sep 1--14 & 218.1 & 200.9 & 378.2 & 4.78 & 4.20 & 7.28 \\
\bottomrule
\end{tabular}
\end{table}

\begin{figure}[ht]
\centering\includegraphics[width=\linewidth]{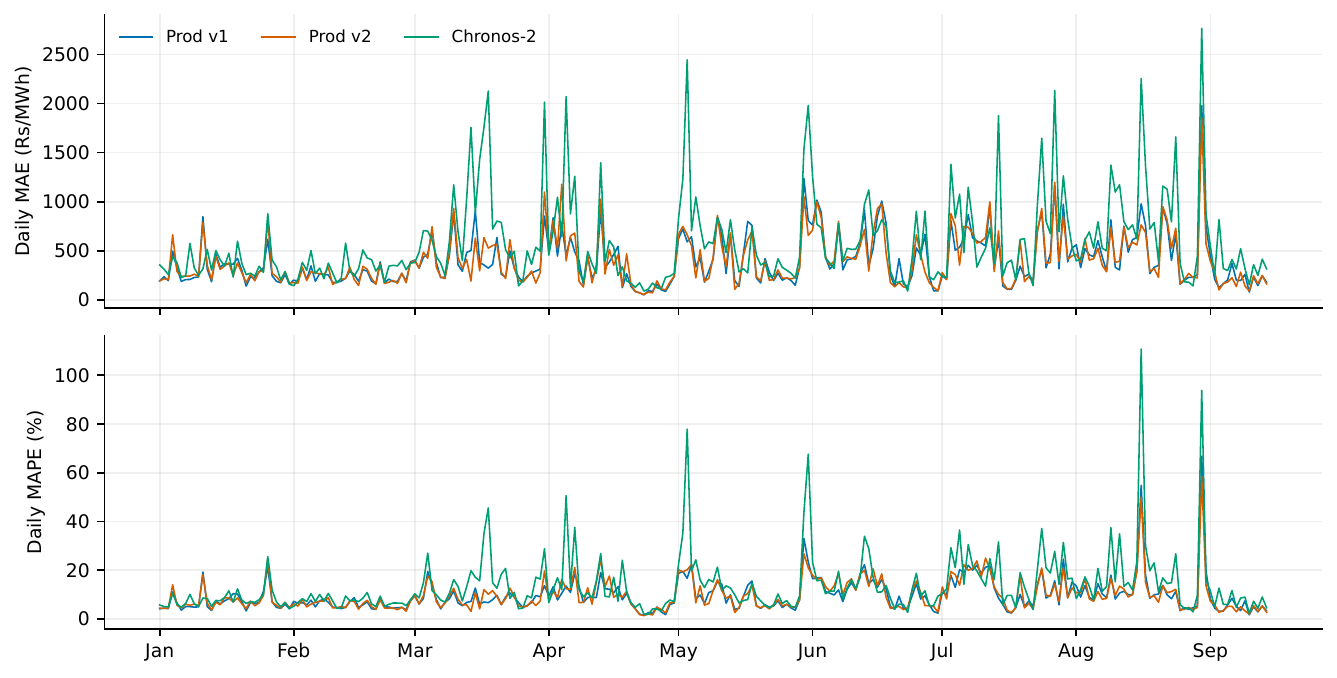}
\caption{Every daily MAE and MAPE value in the 2026 evaluation, without smoothing or removal of difficult days. September ends on day 14.}
\end{figure}
\begin{figure}[ht]
\centering\includegraphics[width=\linewidth]{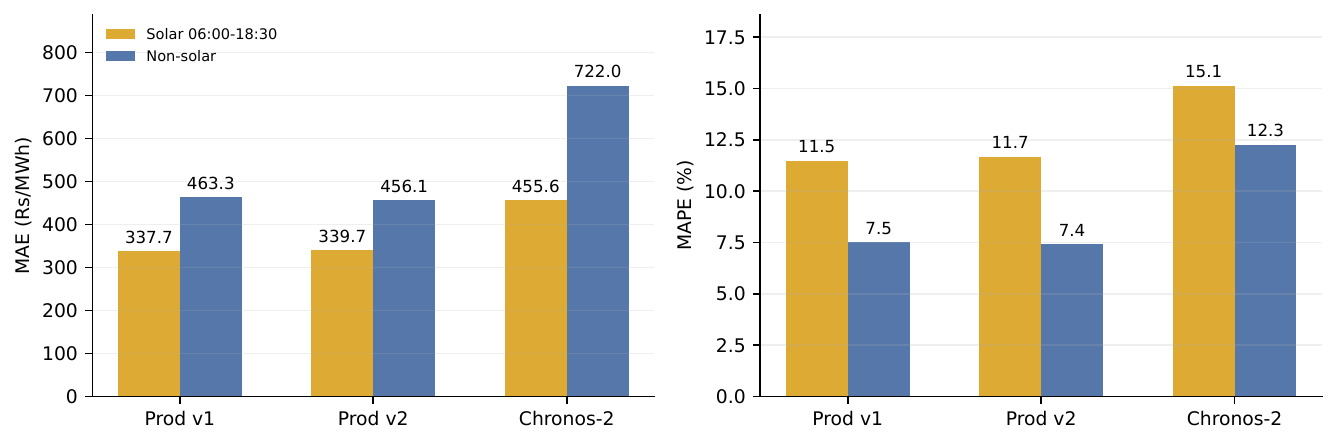}
\caption{Fixed-band solar and non-solar MAE and MAPE for all three saved learned systems. Every day contributes 50 solar and 46 non-solar blocks.}
\end{figure}

\subsection{Block-bootstrap sensitivity}
For a vector of $D=257$ daily loss differences and block length $\ell$, each replicate draws $\lceil D/\ell\rceil$ start indices uniformly from $\{0,\ldots,D-1\}$. Each start yields $\ell$ consecutive indices modulo $D$. The blocks are concatenated, truncated to length $D$, and averaged. Percentile bounds summarize 10,000 replicate means. Circular wrapping avoids giving interior days more resampling opportunities than edge days, while introducing artificial boundary adjacency; this is one reason the intervals are approximate. Large month-to-month distribution changes and only one observed year segment limit their long-run interpretation.
\begin{table}[ht]
\centering\small
\caption{Sensitivity of 95\% paired daily MAE intervals to block length. All differences and bounds are v1 minus comparator in Rs/MWh.}
\label{tab:sensitivity}
\begin{tabular}{lrrrr}
\toprule
Comparator & Block days & $\overline\Delta$ & Lower & Upper \\
\midrule
Prod v2 & 7 & 2.4 & -8.1 & 12.8 \\
Prod v2 & 14 & 2.4 & -7.8 & 12.6 \\
Prod v2 & 28 & 2.4 & -8.6 & 13.2 \\
Chronos-2 & 7 & -185.4 & -244.2 & -133.2 \\
Chronos-2 & 14 & -185.4 & -250.4 & -127.7 \\
Chronos-2 & 28 & -185.4 & -253.9 & -122.9 \\
Previous day & 7 & -191.2 & -237.3 & -146.0 \\
Previous day & 14 & -191.2 & -240.1 & -144.9 \\
Previous day & 28 & -191.2 & -244.6 & -143.0 \\
Previous week & 7 & -843.1 & -1,054.5 & -648.1 \\
Previous week & 14 & -843.1 & -1,086.5 & -620.5 \\
Previous week & 28 & -843.1 & -1,102.7 & -587.7 \\
\bottomrule
\end{tabular}
\end{table}

No interval in this paper is a standard deviation across training seeds. Nor do the resamples simulate new historical market trajectories or remove architecture-selection bias. The sensitivity table tests only one choice in the inference procedure, the temporal block length.

\subsection{Quantile scores and coverage}
Original-price mean pinball loss is
\begin{equation}
 \operatorname{MPL}=\frac{1}{9N}\sum_{i\in\mathcal I}\sum_{q\in\mathcal Q}
 \rho_q\big(y_i-Q_i(q)\big).\label{eq:mpl}
\end{equation}
Coverage of the nominal-80\% interval is $N^{-1}\sum_i\ind\{Q_i(0.1)\leq y_i\leq Q_i(0.9)\}$; its mean width is $N^{-1}\sum_i[Q_i(0.9)-Q_i(0.1)]$. Inclusion is inclusive at both boundaries. The CDF diagnostic uses $N^{-1}\sum_i\ind\{y_i\leq Q_i(q)\}$. With ties and a mass of prices at the upper bound, a correctly specified quantile need not have this empirical proportion exactly equal to $q$. Thus the diagonal in Figure~\ref{fig:uncertainty} is a useful reference, not a test that every discrepancy implies failure.

The reported quantiles are post-calibration outputs. The saved combined files do not provide an unadjusted counterpart for every day, so these scores do not identify the incremental effect of calibration. In particular, the wide v2 intervals cannot be attributed uniquely to its recency weights or any single implementation component from the aggregate saved outputs alone.
\begin{table}[ht]
\centering\small
\caption{Outcome-conditioned diagnostics for low-price and upper-price blocks. MAE and bias are in Rs/MWh; MAPE and nominal-80\% coverage are percentages. The two subsets are disjoint and do not exhaust the evaluation set.}
\label{tab:regimes}
\begin{tabular}{llrrrrr}
\toprule
Actual price & Model & Blocks & MAE & MAPE & Bias & Coverage \\
\midrule
$y<1,000$ & Prod v1 & 464 & 225.6 & 30.60 & 195.0 & 64.01 \\
$y<1,000$ & Prod v2 & 464 & 282.2 & 37.28 & 265.7 & 53.23 \\
$y<1,000$ & Chronos-2 & 464 & 370.5 & 50.96 & 358.1 & 68.10 \\
$y\geq9,990$ & Prod v1 & 6398 & 374.4 & 3.74 & -374.4 & 87.53 \\
$y\geq9,990$ & Prod v2 & 6398 & 344.9 & 3.45 & -344.9 & 94.22 \\
$y\geq9,990$ & Chronos-2 & 6398 & 562.9 & 5.63 & -562.9 & 88.23 \\
\bottomrule
\end{tabular}
\end{table}

The low-price and upper-price subsets are defined after realization. They diagnose performance conditional on market outcome and should not be described as a deployed regime classifier. They also have different sample sizes and days of occurrence. The intermediate price band and high-price totals are retained in the machine-readable analysis for reproducibility.

\section{Pinned later-stack files}
\label{app:future}
Tables~\ref{tab:later}--\ref{tab:later_nov} use:
\begin{itemize}[leftmargin=*,itemsep=2pt]
\item Chronos-2, 25--29 Oct: \path{iex_data_processed/markets/gdam_forecast_daily/gdam{25--29}october.csv}.
\item Chronos-2, 30 Oct--17 Nov: \path{GDAM_ProdV2_InputFix_2026/chronos2_later/gdam_YYYY-MM-DD.csv} (slim checkpoint \texttt{models\_gdam\_chronos2}).
\item exp5 / exp7 dated prediction CSVs under \path{GDAM_ProdV2_InputFix_2026/exp5_*} and \path{exp7_*}, plus the Oct ablation files for 25--28 Oct.
\item InputFix block files under \path{GDAM_ProdV2_InputFix_2026/seasonal_nov_test/nov*_forecasts/}; Oct 25--28 InputFix aggregates from \path{compare_exp5_vs_inputfix/}.
\item Comparison dump: \path{GDAM_ProdV2_InputFix_2026/chronos2_later/paper_compare.json}.
\end{itemize}
17 November has Chronos-2, exp5, and exp7 forecasts but no published actual price at the time of writing, so that day is left out of the scored tables.

\end{document}